\documentclass[conference]{IEEEtran}
\usepackage{blindtext, graphicx, textcomp}
\ifCLASSINFOpdf
\else
\fi
\begin{document}
%
\title{Scale out for large minibatch SGD: \\Residual network training on ImageNet-1K with improved accuracy and reduced time to train}

\author{\IEEEauthorblockN{Valeriu Codreanu}
\IEEEauthorblockA{Compute Services \\
SURFsara B.V. \\
The Netherlands}
\and
\IEEEauthorblockN{Damian Podareanu}
\IEEEauthorblockA{Compute Services \\
SURFsara B.V. \\
The Netherlands}
\and
\IEEEauthorblockN{Vikram Saletore}
\IEEEauthorblockA{AI Products Group \\
Intel Corp.\\
United States}}


%


\maketitle

\begin{abstract}
For the past 5 years, the ILSVRC competition and the ImageNet dataset have attracted a lot of interest from the Computer Vision community, allowing for state-of-the-art accuracy to grow tremendously. This should be credited to the use of deep artificial neural network designs. As these became more complex, the storage, bandwidth, and compute requirements increased. This means that with a non-distributed approach, even when using the most high-density server available, the training process may take weeks, making it prohibitive. Furthermore, as datasets grow, the representation learning potential of deep networks grows as well by using more complex models. This synchronicity triggers a sharp increase in the computational requirements and motivates us to explore the scaling behaviour on petaflop scale supercomputers. In this paper we will describe the challenges and novel solutions needed in order to train ResNet-50 in this large scale environment. We demonstrate above 90\% scaling efficiency and a training time of 28 minutes using up to 104K x86 cores. This is supported by software tools from Intel's ecosystem. Moreover, we show that with regular 90 - 120 epoch train runs we can achieve a top-1 accuracy as high as 77\% for the unmodified ResNet-50 topology. We also introduce the novel Collapsed Ensemble (CE) technique that allows us to obtain a 77.5\% top-1 accuracy, similar to that of a ResNet-152, while training a unmodified ResNet-50 topology for the same fixed training budget. All ResNet-50 models as well as the scripts needed to replicate them will be posted shortly.

\end{abstract}

\begin{IEEEkeywords}
deep learning, scaling, convergence, large minibatch, ensembles.
\end{IEEEkeywords}

%
\IEEEpeerreviewmaketitle

\section{Introduction}

The popularity of deep neural network (DNN) approaches has grown steadily since the, by now famous, AlexNet architecture demonstrated unprecedented levels of performance on the computer vision ImageNet \cite{Russakovsky2015} classification challenge in 2012 \cite{Krizhevsky2012}. This seminal paper paved the way to more and more neural network research, particularly applied to Computer Vision. All subsequent ImageNet challenges have been won by deep neural networks. Due to many neural network training innovations (e.g. residual connections, dropout, batch normalization, etc.) the accuracy on this complex, 1000-category dataset was significantly reduced over the years to a mere 2.25\% top-5 error on the test set by an ensemble of Squeeze and Excitation Networks \cite{Hu2017}. Besides the major impact in this field, end-to-end DNNs have been applied in other fields like voice recognition and natural language processing \cite{LeCun2015} obtaining state-of-the-art accuracies. Recently, Silver \emph{et al.} \cite{Silver2017} obtained a remarkable achievement for the game of Go and improved the previous research performed for the AlphaGo agent. 

These innovations could not have happened if the research community wouldn't have had a computing platform that efficiently supported linear algebra computation, heavily present in deep neural network training. By being able to train neural networks in hours instead of days, researchers could benefit from a shorter research cycle in order to validate their ideas quicker. The single machine approaches have been dominated by NVIDIA GPUs that filled this role for the last 5 years \cite{Schmidhuber2015}. However, as DNN architectures evolve, training "high-accuracy" networks using a single GPU card, or even a single GPU server becomes prohibitive. Presently, some of the most popular DNN designs are based on residual blocks, ever since in 2015 the ImageNet competition was won by He \emph{et al.} \cite{He2016} in 2015. Training a 50 layer residual network (ResNet-50) on the ImageNet-1K dataset takes around 10 days using an NVIDIA P100 GPU card. Training a larger ResNet-152 in the same setting would take roughly more than 3 weeks. As the size and complexity of the training datasets increases, supervised classification systems become even more accurate \cite{Sun2017}. This increases of course the time-to-trained model. By continuing the previous example, it would take roughly one year to train a ResNet-152 model using the full ImageNet-22K dataset, which is about one order of magnitude larger than ImageNet-1K (around 14 million images split across 21841 different image categories).

Decreasing this prohibitive execution time when training complex residual networks on large-scale datasets is the main focus of this work. We believe that this will speed up the research cycle, and will enable researchers outside the deep learning community to adopt DNN techniques in a high performance computing (HPC) environment.

There have been a lot of research around scaling stochastic gradient descent based machine learning algorithms \cite{Dean2012}, \cite{Chilimbi2014}, \cite{Goyal2017}, \cite{Cho2017}, \cite{You2017}, \cite{Chen2016}. Most of these works, as well as our current work focus on data parallel scaling. This effectively means that at each training iteration a large minibatch is evenly divided among the workers. Historically, many algorithms would not scale to large minibatch sizes, as this would either hinder the convergence abilities of the underlying network, or would need many more training epochs in order to reach the desired validation accuracy. Keskar \emph{et al.} \cite{Keskar2016} concluded that large-batch training leads to a  generalization gap, meaning that the trained networks perform poorly on the validation set. This happens because in the large-batch regime training seems to converge to sharp minima of the training and testing functions. The research from \cite{Hoffer2017} suggests that the generalization gap can be partially closed by just training longer. Another recent approach by Dinh \emph{et al.} \cite{Dinh2017} promotes reparametrization as a means to improve the geometry of the minima. On the other hand, Facebook's research \cite{Goyal2017} suggests that large-batch training is more an optimization problem, that can be solved up to a point (global batch size of 8192) using a smarter learning rate schedule (linear LR scaling and gradual warm-up). Another notable example of large-scale training using Intel Knights Landing based systems is the one from \cite{Kurth2017}. In this work the authors scale out deep neural network training up to the size of the full Cori-2 system. However, they do not perform fully synchronous SGD, and although they apply their techniques on various scientific problems, they don't do so on ImageNet.

The experiments presented in this work push large-batch training research forward, both in terms of scaling efficiency and preserving a high validation accuracy for large minibatches of up to 65536 images.

The main contributions of this paper are:
\begin{itemize}
  \item We propose different learning schedules that allow to set new accuracy standards for the unmodified ResNet-50 residual network architecture on the ImageNet-1K dataset, significantly exceeding the result from \cite{He2016}, without using extreme data augmentation.
  \item We scale out synchronous SGD deep learning training to the largest scale to date, reaching convergence on up to 1536 Knights Landing nodes or 1024 Intel Skylake CPUs. We report convergence in 28 minutes using 1536 Knights Landing nodes. We also set a new state-of-the-art accuracy on batch sizes up to 65536.
  \item We propose a collapsing ensemble technique as a reasonable approach for increasing the accuracy of any deep learning model while keeping the training budget constant. Based on this, we observe better ResNet-50 validation accuracy levels for batches up to 16K. We also show the quickest time to each of previously achieved accuracy levels on ImageNet-1K, by using open-source tools from the Intel ecosystem.
\end{itemize}

\section{Background}

In the last year, a lot of research has been conducted on scaling DNNs to a large collection of compute nodes, as the data and network sizes grossly exceeded the compute capabilities of single servers. It is definitely a trend set to continue, as it was suggested already \cite{Sun2017} that if the data neural networks are trained on is expanded, so is the performance achieved by the networks.

Despite the abundance of newly developed methods and techniques, only a few seriously consider the impact of large-batch training on the validation accuracy. Notably, Goyal \emph{et al. }\cite{Goyal2017} is one of the few that report competitive validation accuracy when scaling ResNet-50 training to 32 GPU-enabled servers. Facebook's work is revolutionary, showing excellent scaling properties with the Caffe2 framework, and actually achieving around 1\% better top-1 validation accuracy than the one obtained by the original ResNet authors \cite{He2016}. We adopted the 5 epoch gradual warm-up technique with momentum correction, as well as their proposed batch normalization initialization of the $\gamma$  parameter in our training procedure ($\gamma=0$ for the final BN layer of each residual block).  However, as will be described in \ref{ResNet_training}, we also introduce different techniques in order to overcome the large-batch convergence issues. Also, compared to their work, we perform the ResNet-50 training process at a much larger scale (up to 1536 Intel-based servers versus 32 GPU-based servers in their case), so the pressure on the network interconnect is much higher. 

Another recent related work is the one from IBM, mostly replicating Facebook's result, and using the Caffe and Torch frameworks that are part of their PowerAI DDL. However, although their 50 minutes ResNet-50 training result on ImageNet-1K is faster than Facebook's 60 minutes using comparable amounts of hardware (256 NVIDIA P100 GPUs, split across 64 IBM Power servers). However, the top-1 validation accuracy is around 75\% \cite{Cho2017}, opposed to the 76.3\% obtained by Goyal \emph{et al. } \cite{Goyal2017}. We think that maintaining state-of-the-art accuracy is of paramount importance when scaling up the batch size.

Another related work is the one described in \cite{You2017}. This is similar to our work in terms of scaling and type of hardware used. They introduce the LARS technique to cope with the large-batch training, and similarly to our work they scale the batch size up to 65536. In a recent version of their paper, they have improved the validation accuracy by using data augmentation, similar to our work. The LARS technique seems to help with the optimization difficulties at large batch sizes, and they show they can scale up to 16000 batch size without accuracy loss. We note that we can use a batch size of 32768, and still maintain state-of-the-art accuracy. In the same large-batch training context, recent research from \cite{Smith2017g} suggests that the batch size can be expanded further, up to 65536 for ImageNet-1K, and actually can be increased during training. However, although the authors of \cite{Smith2017g} use a better baseline than the ResNet-50, namely the much heavier Inception-ResNet v2 architecture \cite{Szegedy2017}, they degrade the model performance from above 80\% to below 77\% top-1 accuracy when using large batches. They achieve this result in only 2500 SGD updates. We performed a similar experiment with only 2100 updates but using ResNet-50, and we show we can achieve around 74\% accuracy, less than 2\% degradation compared to ResNet-50's baseline.

Related to these various degrees of validation accuracy levels, Baidu has recently published research around optimizing the mixed-precision training performance of neural networks \cite{Micikevicius2017}. Although their research is not focused on distributed training or scaling, they report a ResNet-50 validation accuracy of 73.75\% when training in mixed-precision, improving from 73.61\% in FP32, both well below state-of-the-art for the ResNet-50 topology.

Although some works mention that they don't use data augmentation, and that this might be the cause for the lower accuracy achieved, we believe that the results from the various papers are not directly comparable, and as opposed to the ones from this paper, are also not reproducible. We thus propose to set some target validation accuracy levels, and report the time needed by the model to reach a given accuracy, such that future comparison can be as fair as possible.

As far as we are aware, besides the research from \cite{Goyal2017} and \cite{You2017}, most the other ResNet-50 scaling works report a validation accuracy lower than the 75.3\% presented in the original paper on residual networks \cite{He2016}. In our view, this also makes the timing results not fully comparable. As we will present in this current work, there is a clear trade-off between the time needed to train a model and the accuracy that is achieved by the model. When training the popular ResNet-50 architecture on ImageNet-1K, the resulting model achieves a validation accuracy of 77\% using a single 224x224 center crop evaluation, by efficiently engineering the training procedure, but without increasing the training budget (number of epochs). Moreover, when using the collapsed ensemble technique, one can reach 77.5\% accuracy, exceeding the ResNet-152 performance, but with the training budget of ResNet-50.  All results can be achieved using servers equipped solely with Intel CPUs.

\section{Distributed Deep Learning using Intel Caffe}

All the experiments performed for this research rely on Intel's optimized branch of the Caffe framework \cite{Caffe2017}. Intel Caffe is quickly following the developments of Caffe's master, but has a clear focus on achieving the highest training performance for Intel architectures. Intel's ML-SL \cite{MLSL2017} (Machine Learning Scaling Library) is a software library that efficiently deals with the communication involved in  neural networks. It is tightly coupled to the training framework, allowing simultaneous compute and communication when performing the backward propagation pass. ML-SL allows for both data and model parallelism, and does that while efficiently using the bandwidth offered by high-speed interconnects such as Intel's OPA. When using model parallelism the model is divided across the participating workers, and workers communicate both in the forward and backward passes. This has the potential to accommodate larger models, as each worker will hold a part of the parameter set. On the other side, in the case of data parallelism, the parameters are replicated on each worker, but communication happens only in the backward pass. We have efficiently scaled synchronous SGD (SSGD) using ML-SL to 1536 Knights Landing nodes (\textasciitilde 104K x86 cores). In this work we will only focus on the data parallel case, as we believe current hardware is more efficiently used in this manner, especially since CPU-based servers feature enough memory to hold even the largest models.

\section{Experimental methodology}

In this section we describe the changes performed to the typical training procedure of ResNet-50. We emphasise the techniques that allow for large batch training.

\subsection{ImageNet-1K dataset}

We evaluate our method on the ImageNet 2012 classification dataset \cite{Russakovsky2015} that consists of 1.28 million training images split across 1000 classes. All our models are trained on the full 1.28 million images, and evaluated on the 50,000 validation images from the standard ILSVRC2012 validation set.

\subsection{ResNet-50 training methodology}
\label{ResNet_training}

The goal of our research is the fastest time to a given validation accuracy level. We worked on achieving this goal using the ResNet-50 architecture, and altered parts of the training methodology. These are described below. We note that by using this training procedure our models can achieve validation accuracy much higher than the ones presented by \cite{He2016}, peaking at around 77\% top-1 accuracy after 120 epochs. As noted before, we use all the 50,000 images from the ILSVRC2012 validation set for evaluating the performance of our models. As for data augmentation, we only employ basic scale and aspect ratio augmentation. All model evaluations were done on a single 224x224 center crop extracted from the 256x256 resized input image, unless otherwise noted. Also, our models were trained from scratch. We used standard values for the hyperparameters. The momentum value was set to 0.9, and the weight decay, $\lambda$, to 0.0001, as these values seem to work quite well up to a relatively large batch size. To push the performance of the model further, we change the weight decay value dynamically during training as will be further explained in Section \ref{convergence}.

\subsection{Batch normalization considerations}

Before starting the experiments on ImageNet-1K using the ResNet-50 architecture, we performed an extensive set of experiments using the Inception-v1 architecture, that does not feature batch normalization layers \cite{Ioffe2015}. It was noticeable that this was limiting the model performance when using a large-batch for training. During training we used normalization over the current minibatch and global statistics were accumulated by a moving average. During testing, the accumulated mean and variance values were used for normalization. Smaller values make the moving average decay faster, giving more weight to the recent values. With each iteration the moving average is updated with the current mean becoming $S_t = (1-\beta)Y_t + \beta \cdot S_{t-1}$, where $\beta$ is the moving average fraction parameter. For the moving average fraction, we have empirically set the value of the parameter from Caffe\textquotesingle s BatchNorm layer to 0.95 for all BN layers.

\subsection{Learning rate schedule}
\label{learning_rate}

The most widely used method to decay the learning rate is the classical 3-step 10-fold decrease. Empirically, we noticed that it is more effective to use polynomial decay with the power of 1 for the learning rate decay schedule (basically linear decay, same as the linear increase from the warm up phase), instead of the classic 3-step decrease. To be sure that this is the case we have performed some ablation experiments. An example comparative experiment for a 90 epoch training run using 240 nodes is presented in Table \ref{table_poly_step}. For this experiment we kept all settings fixed besides changing the learning rate from a 3-step decay (at epoch 30, 60, and 80) to a linear decay. We note that both runs include a warm-up phase of 5 epochs out of the 90 epoch training budget, as described in \cite{Goyal2017}, and the value from which the learning rate decreases follows the linear scaling rule proposed in the same work. All further experiments in the paper use a polynomial decrease of the LR.

\begin{table}[!t]
\renewcommand{\arraystretch}{1.3}
\caption{3-step decay versus linear decay of the learning rate}
\label{table_poly_step}
\centering
\begin{tabular}{|c|c|c|c|c|c|}
\hline
Batch size & \# nodes & \# epochs & accuracy [\%] \\
\hline
7680 & 240 & 90 & 75.68/92.95  \\
\hline
7680 & 240 & 90 & 75.44/92.69 \\
\hline
\end{tabular}
\end{table}

When using a polynomial decay, the learning rate decreases from its original value to 0 over the number of training iterations. This allows us to easily control the duration of the training, the only detail changing between a quick run an a full run being the decay slope of the learning rate. Up to a global batch size of 8K/16K we noticed that if we aggressively decrease the learning rate, we can achieve good model performance even faster into the training (quicker than 90 epochs). To achieve the last bit of performance we perform the final 3-5 epochs in a "collapsed" fashion with augmentation disabled, as will be explained in Section \ref{collapsed_ens}. We therefore set four performance levels for ResNet-50, namely:

\begin{itemize}
  \item 75.5\% top-1 accuracy achieved using 48 training epochs. The results from the original ResNet implementation \cite{He2016} falls in this category.
  \item 76\% top-1 accuracy achieved using 64 training epochs. The result from \cite{Goyal2017} falls in this category.
  \item 76.5\% top-1 accuracy achieved using 78 training epochs.  This is a state-of-the-art result for large-batch training.
  \item a new SOTA (state-of-the-art) 77\% top-1 accuracy when training for 120 training epochs.

\end{itemize}

Sections \ref{results} and \ref{collapsed_ens} will describe both scaling and convergence results in more detail. 

All our results also use the warm-up scheme proposed in \cite{Goyal2017}, and failing to use it can lead to noticeably higher validation error levels.

\subsection{Execution methodology}

All experiments were executed on either Intel Knights Landing based servers, or dual-socket Intel Skylake servers. We use three separate HPC infrastructures for this study:

\begin{itemize}
  \item TACC's Stampede2 system composed of Intel Knights Landing 7250 nodes, each with 96GB of RAM. All experiments are performed with the nodes set in Cache-Quadrant mode. We use this system to perform the KNL scaling experiments.
  \item BSC's MareNostrum4 system composed of dual-socket Skylake 8160 nodes, each with 96GB of RAM. We use this system to perform the Skylake scaling experiments.
  \item An internal Intel Knights-Landing cluster with 192GB of RAM per node, and large local storage. The KNLs in this system are configured in Flat-Quadrant mode. We use this system for ablation studies and for training exploration.

\end{itemize}

When using distributed training with Intel Caffe and Intel ML-SL, it is needed to carefully choose the number (and IDs) of cores that are performing communication and computation. In the case of the KNL experiments, the affinity is set explicitly so that the ML-SL processes use the last 4 cores to perform communication. The first 64 cores from the core list take part in the OpenMP team for computation. In the Skylake case, since it is a dual socket system, we have noticed that we achieve the best performance by pinning 2 EP servers on each of the sockets, and 22 OpenMP threads on each socket, for a total of 44 OpenMP threads participating in the computation (the Skylake 8160 part features 24 cores per socket).

\section{Experimental results}
\label{results}

\subsection{Scaling ResNet-50 training up to 12K batch size}

Analogously to the large-batch training results of Facebook and IBM, where they have scaled up to 256 NVIDIA P100 GPUs, we also scale the training procedure of ResNet-50 on up to 256 Intel Knights Landing and Intel Skylake nodes. In order to be as close as possible in terms of methodology, we also employ a minibatch size of 32 per worker. However, it is clear that scaling to 256 separate nodes puts much greater pressure on the communication interconnect compared with scaling to 32 Big Basin nodes in the case of Facebook or 64 Minsky nodes in the case of IBM. Moreover, as will be presented in Section \ref{largest_batch}, we do not hit a limit at 256 nodes, and scale ResNet-50 training all the way up to 1536 KNL nodes as well as on up to 1024 SKX CPUs. All our scaling experiments are performed on production systems such as TACC's Stampede Intel Xeon Phi-based supercomputer as well as BSC's MareNostrum 4 Intel Skylake-based supercomputer.

The duration of a training epoch is a function of the local batch size and number and types of workers participating in the run. In our fastest scenario we can go through the 1.28 million training images from ILSVRC2012 in around 16.8 seconds while using 1536 KNL nodes, each working with a local batch size of 32 examples. This is the largest batch size (49152) we have experimented so far with on a production system. However, for this very large-batch scenario, at the moment our ResNet-50 needs around 100 epochs to reach a reasonable accuracy of 74.6\%. The full training run is performed in 28 minutes, as will be described in \ref{largest_batch}. This is significantly faster than the 33.3 seconds from \cite{Cho2017}, and 40 seconds from \cite{Goyal2017} for processing one epoch on ImageNet-1K using the ResNet-50 architecture. You \emph{et al.} \cite{You2017} can process one epoch in 20.66 seconds in the latest version of their paper, while using 1600 Intel Skylake CPUs. 

The results presented in this section fall in two categories: experiments using the full 90 epoch training schedule, experiments with less than 90 trained epochs showing the trade-off between final accuracy and training time.

\subsubsection{TACC Stampede2 results}

The KNL scaling experiments were performed on the Stampede2 system at TACC. Stampede2 features 4200 Xeon Phi 7250 nodes connected with Intel OPA fabric, each node being provided with 96GB RAM (HBM configured in cache mode). Since we are concerned with ImageNet-1K, that is represented by a 42GB compressed LMDB file, we could easily copy the dataset to RAM at the beginning of the job, this clearly improving execution efficiency. 

We discovered empirically that after 5 warm-up epochs, and around 24 training epochs following the linear learning rate decay policy as described in Section \ref{learning_rate}, a ResNet-50 model can achieve around 73\% accuracy. After around 50 epochs of training, the model achieves 74\%. At the other extreme, after 90 epochs of training, all our models achieve 75-75.8\% top-1 validation accuracy.

Table \ref{table_stampede2_sota} presents the validation accuracy results obtained on ImageNet-1K dataset using single-crop evaluation of ResNet-50 models, as well as the total time-to-train required to achieve it. We train the exact same model, with the same training strategy as in the previous subsections; the only parameter changed being the number of training epochs. Performing a full 90 epoch schedule leads to a SOTA top-1 validation error on the full ILSVRC2012 validation set. Albeit the runtime is longer due to the actual 90-epoch run training budget as well as smaller local batch sizes (16 in the 720 node case), we show that our training architecture is capable of delivering SOTA accuracy on ResNet-50. We note that our 75.81\% accuracy is actually higher than the 75.3\% obtained by the ResNet authors (although we use a much higher batch size), and also higher than the one presented by \cite{Cho2017}. 

\begin{table}[!t]
\renewcommand{\arraystretch}{1.3}
\caption{ResNet-50 SOTA results on Stampede2}
\label{table_stampede2_sota}
\centering
\begin{tabular}{|c|c|c|c|c|}
\hline
Batch size & \# nodes & \# iterations (epochs) & accuracy [\%] & TTT[min] \\
\hline
8192 & 256 & 14200 (90) & 75.81/92.93 & 140 \\
\hline
12288 & 512 & 9400 (90) & 75.25/92.90 & 80 \\
\hline
11520 & 720 & 10000 (90) & 75.03/92.69 & 62 \\
\hline
\end{tabular}
\end{table}

One of our goals was to decrease the time to train for a good quality model. Thus, we've set an accuracy target of 74\% top-1, and reduced the number of epochs in order to reach this target, as quick as possible. In Table \ref{table_stampede2_74} we present results that were obtained using a various number of training epochs, batch sizes, and number of nodes, all reaching above 74\% validation accuracy.  The results show an accuracy level of 74.05/92.07 top-1/top-5 for the 512-node case within 46 minutes, and 74.20/92.20 for the 768-node case in only 39 minutes. This is achieved while only using a local minibatch size of 16, as there are clear optimization difficulties above 12K global minibatch sizes. The means to overcome them are discussed in Section \ref{largest_batch}. Perhaps an interesting observation from these experimental results obtained on Stampede2 is that in the case of 256-node run using the smaller local minibatch size of 16 (first row from Table \ref{table_stampede2_74}), we can achieve the target validation accuracy by only training for 37 epochs. This result is very close to the Facebook/IBM ones in terms of time-to-train, achieving convergence at above 74\% top-1 accuracy in just 63 minutes while using 256 KNL nodes. 

As in the previous experiments, there are no additional modifications to the setup, except for the aforementioned standard data augmentation (scale and aspect ratio augmentation), linear scaling of the LR, warm-up, and the described changes in the batch normalization layers.

\begin{table}[!t]
\renewcommand{\arraystretch}{1.3}
\caption{Stampede2 fastest-time-to-74\%}
\label{table_stampede2_74}
\centering
\begin{tabular}{|c|c|c|c|c|}
\hline
Batch size & \# nodes & \# iterations (epochs) & accuracy [\%] & TTT[min] \\
\hline
4096 & 256 & 11560 (37) & 74.05/92.12 & 63 \\
\hline
8192 & 256 & 7800 (50) & 74.12/92.16 & 70 \\
\hline
8192 & 512 & 7800 (50) & 74.12/92.16 & 49 \\
\hline
12288 & 512 & 5600 (54) & 74.05/92.07 & 46 \\
\hline
12288 & 768 & 11560 (37) & 74.20/92.20 & 39 \\
\hline
\end{tabular}
\end{table}

Ultimately, we show that we can maintain competitive accuracy while decreasing the number of training epochs even further. The results from Table \ref{table_stampede2_quick} indicate that one can trade off validation accuracy by training a given network for a lower number of epochs. We can maintain an above 73\% validation accuracy by only training for around 30-40 epochs. Time-wise, we can train ResNet-50 on ImageNet-1K using 256 nodes in only about 40 minutes. We note that one KNL node can only achieve around half of the peak performance of NVIDIA P100 GPUs.

\begin{table}[!t]
\renewcommand{\arraystretch}{1.3}
\caption{ResNet-50 fastest time results on Stampede2}
\label{table_stampede2_quick}
\centering
\begin{tabular}{|c|c|c|c|c|}
\hline
Batch size & \# nodes & \# iterations (epochs) & accuracy [\%] & TTT[min] \\
\hline
2048 & 256 & 18120 (29) & 72.43/91.45 & 64 \\
\hline
4096 & 256 & 11560 (37) & 74.05/92.12 & 63 \\
\hline
8192 & 256 & 4530 (29) & 73.21/91.58 & 41.5 \\
\hline
8192 & 256 & 5780 (37) & 73.92/91.83 & 52 \\
\hline
8192 & 512 & 5780 (37) & 73.78/91.78 & 37 \\
\hline
\end{tabular}
\end{table}

\begin{figure}[!t]
\centering
\includegraphics[width=3.5in]{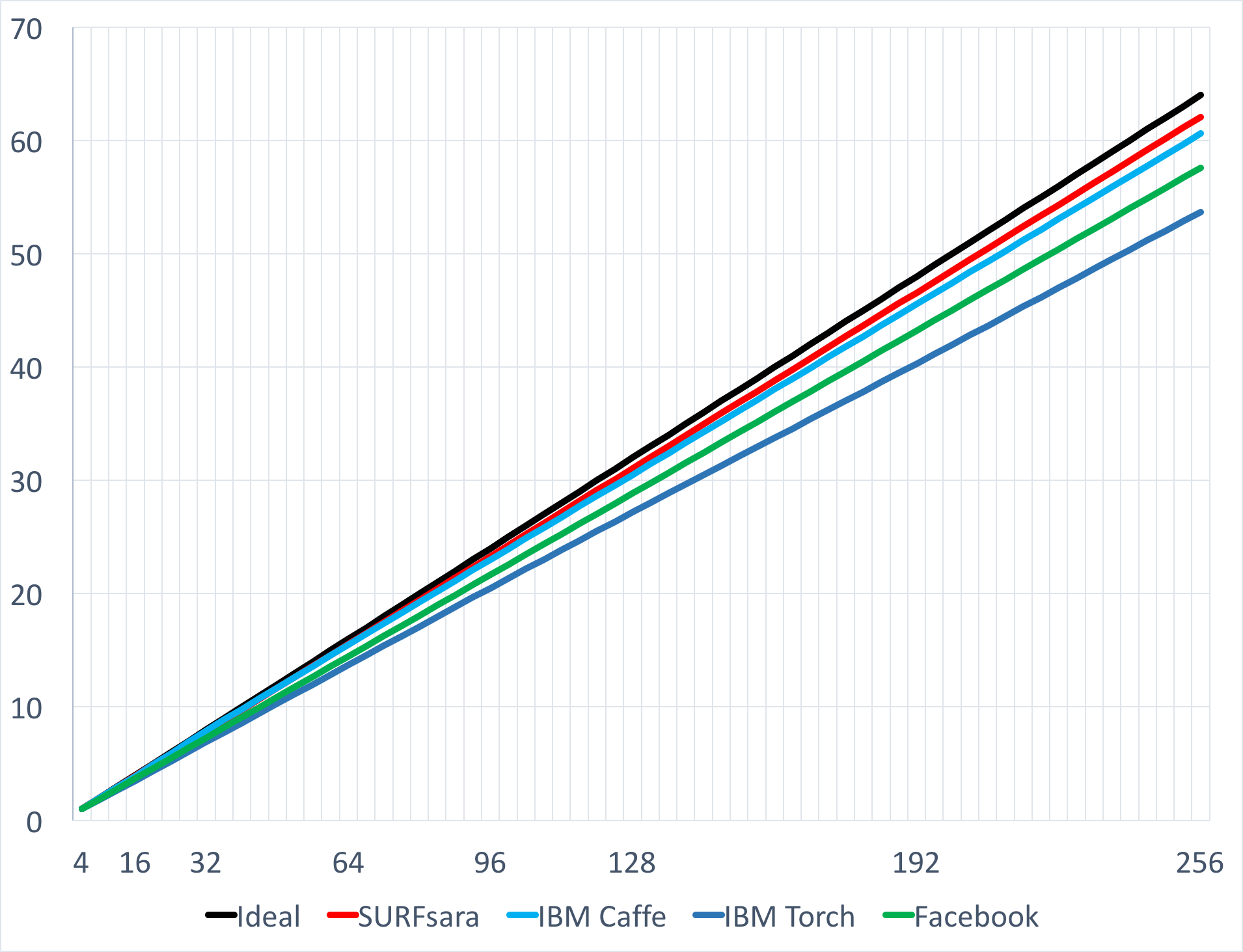}
\caption{Scaling efficiency on Stampede2 (speedup vs number of workers). This plot starts from scaling on 4 workers, which has a scaling factor of 1}
\label{fig_stampede_scaling}
\end{figure}

When taking scaling efficiency into consideration, our approach based on Intel Caffe and Intel ML-SL can achieve an even higher value (above 97\%) when going from 1 to 256 Knights Landing nodes compared to \cite{Goyal2017} and \cite{Cho2017}. This is supported by Figure \ref{fig_stampede_scaling}. A detail to consider here is that our approach uses 4 times the amount of nodes that IBM used, and 8 times the amount Facebook used, leading to an increased amount of data that has to be transferred among the workers. Basically, after each forward pass, each worker node needs to send and receive a complete model (around 98MB for ResNet-50). Thus, in our case, after each iteration 256 worker nodes send and receive 98MB, leading to 49GB of traffic. In the case of IBM this pressure is only 12GB, while in the Facebook case it is only 6GB, due to the intra-node allreduce achieved through NVIDIA's NCCL library.

In order to see how well this efficient scaling holds, we decided to verify some of the strong scaling properties and push the training of ResNet-50 to 512 nodes and above. We attempted 512, 768, and 1024 Xeon Phi node runs with a batch size of only 16 images per node. This led to a global batch size of 16384 for the 1024-node case. Of course this imposes more communication compared to the cases that use a local batch size of 32, as the gradient transfers are performed more frequently in the case of small local batches since computation takes less time. We have also pushed the weak scaling experiments further, scaling the 32 local batch size per worker case to 1536 KNL nodes. We achieve 74.6\% top-1 accuracy in this scenario after just 2600 SSGD updates, representing around 100 epochs, as will be discussed in Section \ref{convergence}. The scaling properties for 1024 and 1536 node runs are still excellent considering the amount of communication involved. The weak scaling efficiency for a local batch size of 16 is 81\% when going from 256 to 1024 nodes. When using a local batch size of 32, the weak scaling efficiency is 88\% when going from 256 to 1024 KNL workers, and above 81\% when going from 1 to 1536 KNL nodes.

Table \ref{time_per_epoch} outlines the number of seconds required for training one full epoch of ImageNet-1K under the various batch sizes and number of nodes. The time-per-epoch depends on the local batch size (more or less efficient use of the hardware) and the number and types of participating nodes (more or less communication). Each value is calculated as an average over a full training run.

The techniques developed subsequently and described in \ref{convergence} or the Collapsed Ensemble techniques from Section \ref{collapsed_ens} are not used in any of the experiments presented in this subsection, so these results can certainly be improved. 

\begin{table}[!t]
\renewcommand{\arraystretch}{1.3}
\caption{Time per epoch on Stampede2}
\label{time_per_epoch}
\centering
\begin{tabular}{|c|c|c|}
\hline
Local batch size & \# nodes & time per epoch [sec] 
\\
\hline
16 & 256 & 102 \\
\hline
32 & 256 & 85 \\
\hline
16 & 512 & 60 \\
\hline
20 & 512 & 55 \\
\hline
24 & 512 & 51.3 \\
\hline
32 & 512 & 48.8 \\
\hline
16 & 768 & 38 \\
\hline
16 & 1024 & 31.5 \\
\hline
32 & 1024 & 24.5 \\
\hline
32 & 1536 & 16.8 \\
\hline
\end{tabular}
\end{table}

%
%

\subsubsection{BSC MareNostrum 4 results}
In parallel with performing the above experiments on Stampede2, we have also performed similar ImageNet-1K experiments on Intel Xeon Skylake nodes on the MareNostrum 4 supercomputer from the Barcelona Supercomputing Center (BSC). It has around 4000 2S Intel Xeon Processor 8160 (SKX) nodes, each featuring 96GB RAM and around 200GB of local storage. One of the original project goals was to evaluate deep learning training on traditional Xeon processors, as these form the backbone of most computing centers around the world. Thus, we chose the versatile Skylake architecture for this purpose.

\begin{figure}[!t]
\centering
\includegraphics[width=3.5in]{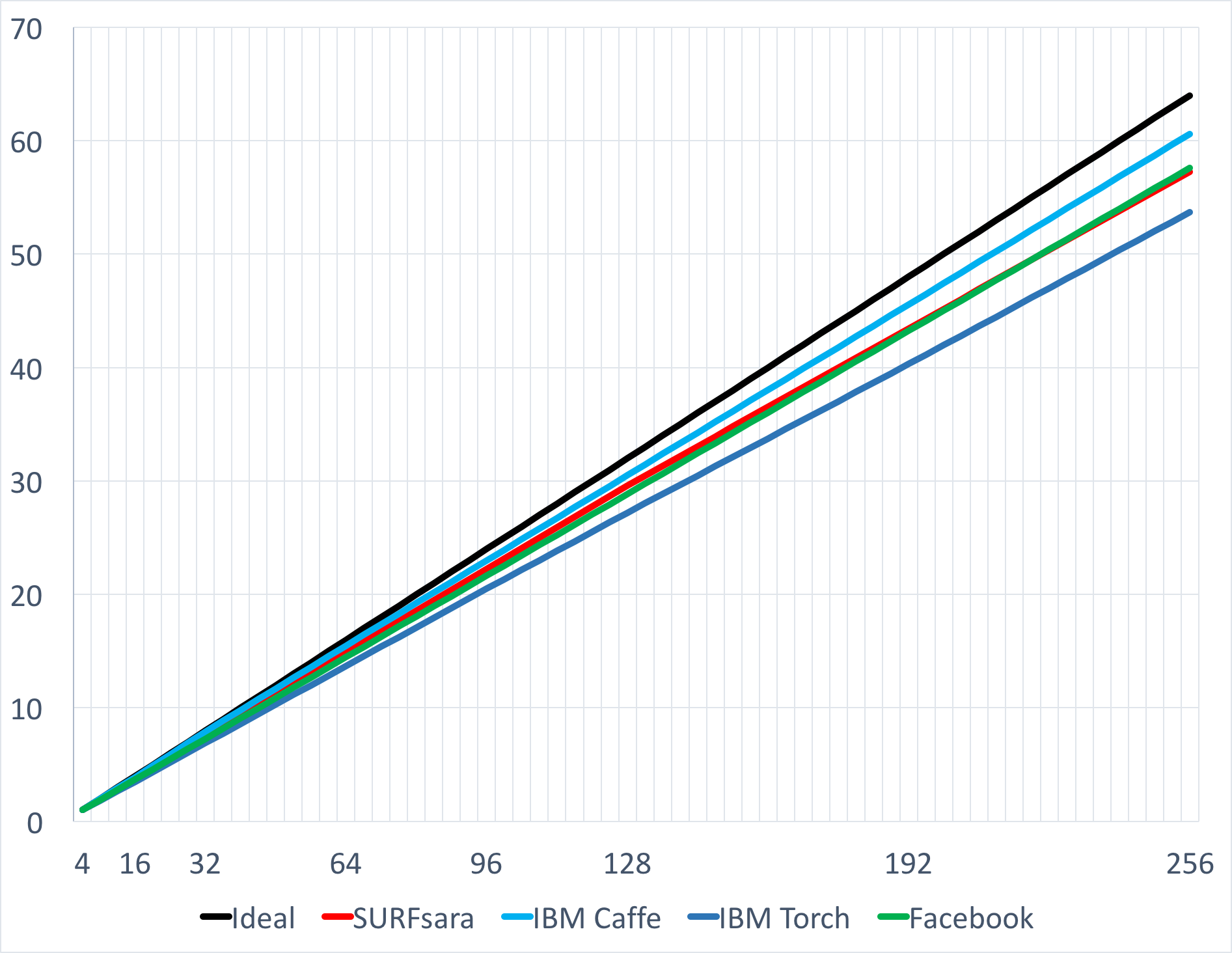}
\caption{Scaling efficiency on MareNostrum 4 (speedup vs number of workers). This plot starts from scaling on 4 workers, which has a scaling factor of 1}
\label{fig_mn_scaling}
\end{figure}

The results were quite successful, leading to around 90\% scaling efficiency when going from 1 to 512 CPUs, as can be seen in Figure \ref{fig_mn_scaling}. Larger runs (512+ CPUs) together with their validation accuracies and achieved time-to-train are presented in Table \ref{table_marenostrum_74}. As in the Stampede2  case, we have set a target accuracy for all our trained models to be greater than 74\% on ILSVRC2012 validation set. These runs are very useful to also compute the throughput, and ultimately the time-per-epoch. What is interesting to notice is that for strong-scaling purposes the Skylake architecture appears to be more appropriate. This is suggested by the fact that the 512-node, 16 batch size per node experiment performed quicker than on the KNL-based Stampede2 system using a similar number of nodes. Due to our limited compute budget we could not perform larger scale experiments (1024+ CPU runs) on MareNostrum4, but this is certainly something we will follow up on. We also did not perform full 90 epoch experiments on MareNostrum 4, but the conclusions from Stampede2 large-batch experiments obviously hold for this architecture too. Even so, these results demonstrate that general-purpose Intel Xeon-based CPU system can perform deep learning training on par with state-of-the-art GPU systems.

\begin{table}[!t]
\renewcommand{\arraystretch}{1.3}
\caption{Time per epoch on MareNostrum 4}
\label{time_per_epoch_mn4}
\centering
\begin{tabular}{|c|c|c|c|}
\hline
Local batch size & \# CPUs & Average throughput & time per epoch [sec] 
\\
\hline
32 & 512 & 15170 &  84.3 \\
\hline
24 & 800 & 20480 & 62.5 \\
\hline
16 & 800 & 19047 &  67.2 \\
\hline
16 & 1024 & 24380 & 52.5 \\
\hline
\end{tabular}
\end{table}

Table \ref{time_per_epoch_mn4} gives the average time per epoch to perform ResNet-50 training using the Intel Skylake architecture in various configurations. The measured runs from Table \ref{table_marenostrum_74} are used as a basis to compute the time per epoch.

\begin{table}[!t]
\renewcommand{\arraystretch}{1.3}
\caption{MareNostrum 4 fastest-time-to-74\%}
\label{table_marenostrum_74}
\centering
\begin{tabular}{|c|c|c|c|c|}
\hline
Batch size & \# CPUs & \# iterations (epochs) & accuracy [\%] & TTT[min] \\
\hline
8192 & 512 & 7800 (50) & 74.11/92.14 & 70 \\
\hline
9600 & 800 & 8000 (60) & 74.29/92.29 & 62.5 \\
\hline
6400 & 800 & 10000 (50) & 73.94/92.03 & 56 \\
\hline
6400 & 800 & 15000 (75) & 74.85/92.58 & 80 \\
\hline
8192 & 1024 & 10000 (64) & 74.15/92.25 & 56 \\
\hline
8192 & 1024 & 7800 (50) & 74.01/92.03 & 44 \\
\hline
\end{tabular}
\end{table}

Using 74\% as our top-1 performance target, we list below some of the highlights of our results:
\begin{itemize}
  \item convergence in 70 minutes using 512 SKX CPUs
  \item convergence in 56 minutes using 800 SKX CPUs
  \item convergence in 44 minutes using 1024 SKX CPUs
\end{itemize}

We again note that the peak performance of the hardware used by Facebook and IBM in their experiments is more than double compared to 512 SKX CPUs, and thus our results counter-balance this through both scaling efficiency as well as training efficiency by employing much quicker learning schedules with fine-grain control over the obtained accuracy.

\subsection{Improving the baseline accuracy}
\label{convergence}

When performing the various scaling experiments, we have clearly noticed that the validation accuracy degrades after exceeding a batch size of around 8K. We discuss here the ways we found that can overcome this behaviour, and also allow for state-of-the-art accuracy on all batch sizes for the case of ResNet-50.

\subsubsection{Smaller weight decay}

We have empirically noticed that the weight decay hyperparameter has a big impact on the optimization difficulties, particularly in the first phase of training, when the learning rates tend to be large. We have empirically noticed that we can improve the final model accuracy by performing the training run identically as before, but with a smaller value for the weight decay hyperparameter. We have chosen half of its default value, 0.00005. By just using this, we can improve our model accuracy from 75.8\% top-1 accuracy to 76.22\% top-1 accuracy.

\subsubsection{Final collapse}

Another empiric observation is that the weight decay needs to actually be dynamically adjusted. We train the first and largest chunk of the training with the lower weight decay and still with a linear learning rate decay. Afterwards, the last 5-7\% of the training is performed in a collapsed fashion. The learning rate is decayed with a power of 2 polynomial in this phase (starting from the value where it left off), the weight decay doubled to 0.0001, and scale/aspect ratio augmentation is disabled. This regime is typically performed for 4-7 epochs. By using this technique we can increase the top-1 accuracy to 76.65\%. This is the best ResNet-50 result we are aware of, and it can be actually achieved at various batch sizes (up to 16K).

The results using this improvements are presented in Table \ref{table_collapse_wd}. They are performed on the VLAB cluster using 128 KNL nodes and a batch size of 4096. These two techniques hold to much larger batch sizes, as is presented in the following section.

\begin{table}[!t]
\renewcommand{\arraystretch}{1.3}
\caption{Impact of techniques to improve baseline accuracy at large batch sizes}
\label{table_collapse_wd}
\centering
\begin{tabular}{|c|c|c|c|}
\hline
Batch size & weight decay ($\lambda$) & collapse & accuracy [\%] \\
\hline
4096  & 0.0001 & no & 75.8 \\
\hline
4096 &  0.00005 & no & 76.22 \\
\hline
4096 &  0.0005+0.0001 & yes & 76.65 \\
\hline
\end{tabular}
\end{table}

\subsection{Scaling to extremely large batch sizes}
\label{largest_batch}

Besides offering large accuracy gains on batch sizes in the order of 4096-8192, these techniques help generalization for batches of up to 64K with (close to) state-of-the-art accuracy. However, for some of these largest runs we do not scale the learning rate linearly. We keep it at a value of 6.4 (for all runs with batch size 16K and above) and decrease it in the same linear way. Also, the warm-up period is typically longer, at around 7-10 epochs. Even with these techniques accuracy degrades as the batch size increases over a certain limit.

We achieve 74.6\% accuracy using a batch size of 49152 distributed over 1536 KNL nodes in just 28 minutes. We also achieve state-of-the-art 75.3\% accuracy at a batch size of 32K, and 74\% at a batch size of 64K in just 2100 iterations, sacrificing 1.3\% top-1 accuracy compared to the original implementation \cite{He2016}. The large batch results are present in Table \ref{table_large_batch}.

\begin{table}[!t]
\renewcommand{\arraystretch}{1.3}
\caption{Very large batch training results}
\label{table_large_batch}
\centering
\begin{tabular}{|c|c|c|c|c|c|}
\hline
Batch size & \# nodes & \# iterations (epochs) & accuracy [\%] & TTT[min] \\
\hline
10240 & 512 & 11400 (91) & 76.40/93.16 & 82 \\
\hline
16384 & 512 & 7200 (92) & 76.26/93.19 & 74 \\
\hline
32768 & 1024 & 4000 (100) & 75.31/92.70 & 42 \\
\hline
49152 & 1536 & 2600 (100) & 74.6/92.1 & \textbf{28} \\
\hline
65536 & 128 & 2100 (107) & 73.94/91.75 & N/A \\
\hline
\end{tabular}
\end{table}

\begin{figure}[!t]
\centering
\includegraphics[width=3.5in]{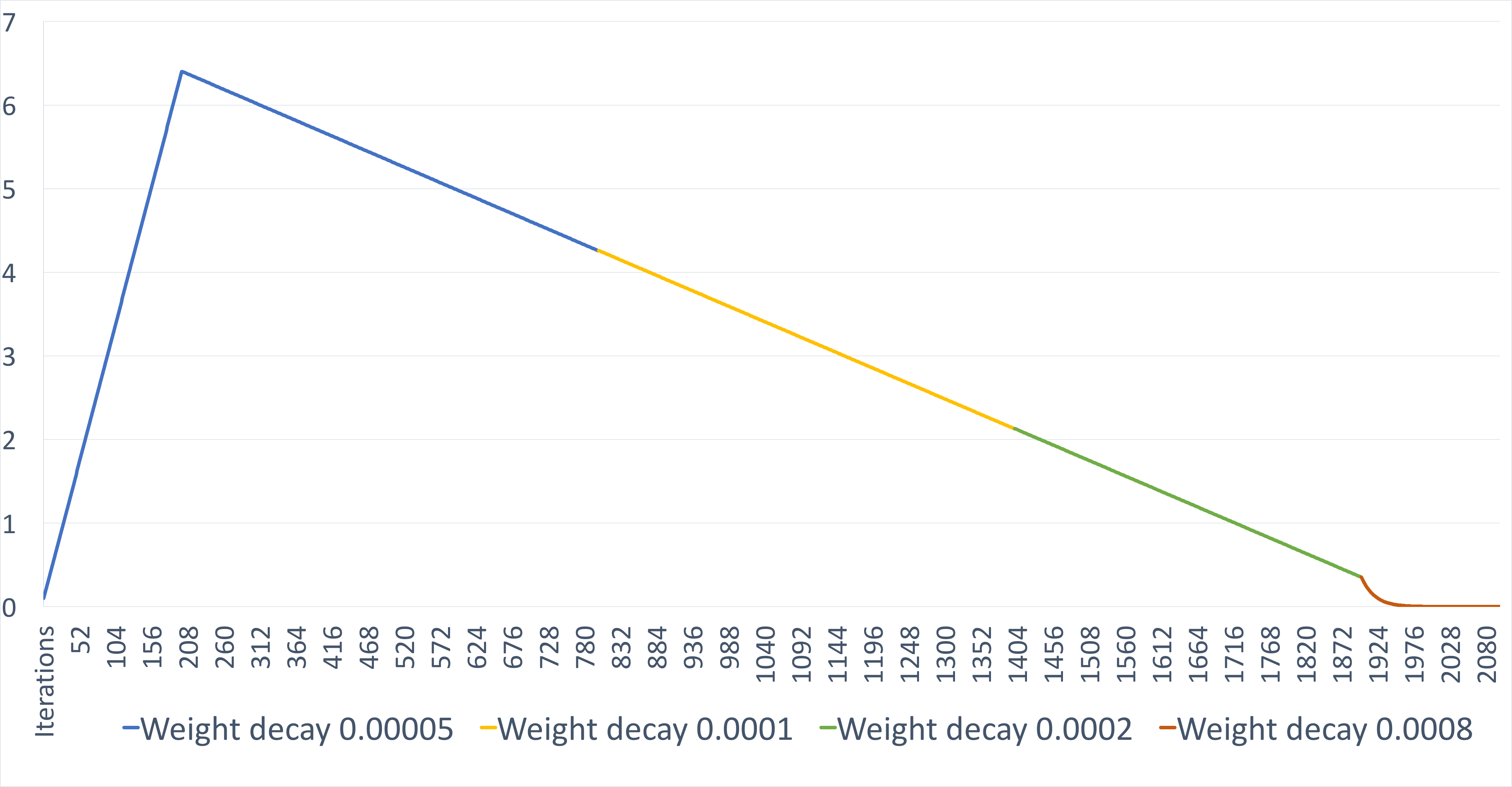}
\caption{Plot of learning rate behaviour when performing a final collapse and coloured for different weight decays for 64k batch size}
\label{fig_collapse}
\end{figure}

The 65536 batch size experiment was performed on the VLAB infrastructure using 128 KNL nodes, with a local batch size of 512. To reach this level of accuracy we have used an even more different learning schedule that we will certainly explore further. The schedule is presented in Figure \ref{fig_collapse} and explores multi-step weight decay increase during training.

Table \ref{table_comparison} compares our results to the most relevant approaches from the literature that consider very large batch training, and show that we exceed the state-of-the-art accuracy for every batch size, while also maintaining very good hardware and scaling efficiency.

\begin{table}[!t]
\renewcommand{\arraystretch}{1.3}
\caption{Comparative accuracy of large batch training. Original ResNet-50 accuracy is 75.3\% top-1}
\label{table_comparison}
\centering
\begin{tabular}{|c|c|c|c|c|c|c|}
\hline
Batch size & 8K & 16K  & 32K & 48K & 64K \\
\hline
\hline
IBM \cite{Cho2017} & 75\% & - & - & - & - \\
\hline
Facebook \cite{Goyal2017} & 76.2\% & 75.2\% & 72.4\% & - & 66.04\% \\
\hline
You \emph{et al.} \cite{You2017}  & 75.3\% & 75.3\% & 74.7\% & - & 72\% \\
\hline
This work & \textbf{76.6\%} & \textbf{76.26\%}  & \textbf{75.31\%} & \textbf{74.6\%} & \textbf{73.94\%} \\
\hline
\end{tabular}
\end{table}

\section{Collapsed ensembles}
\label{collapsed_ens}
\begin{figure}[!t]
\centering
\includegraphics[width=3.5in]{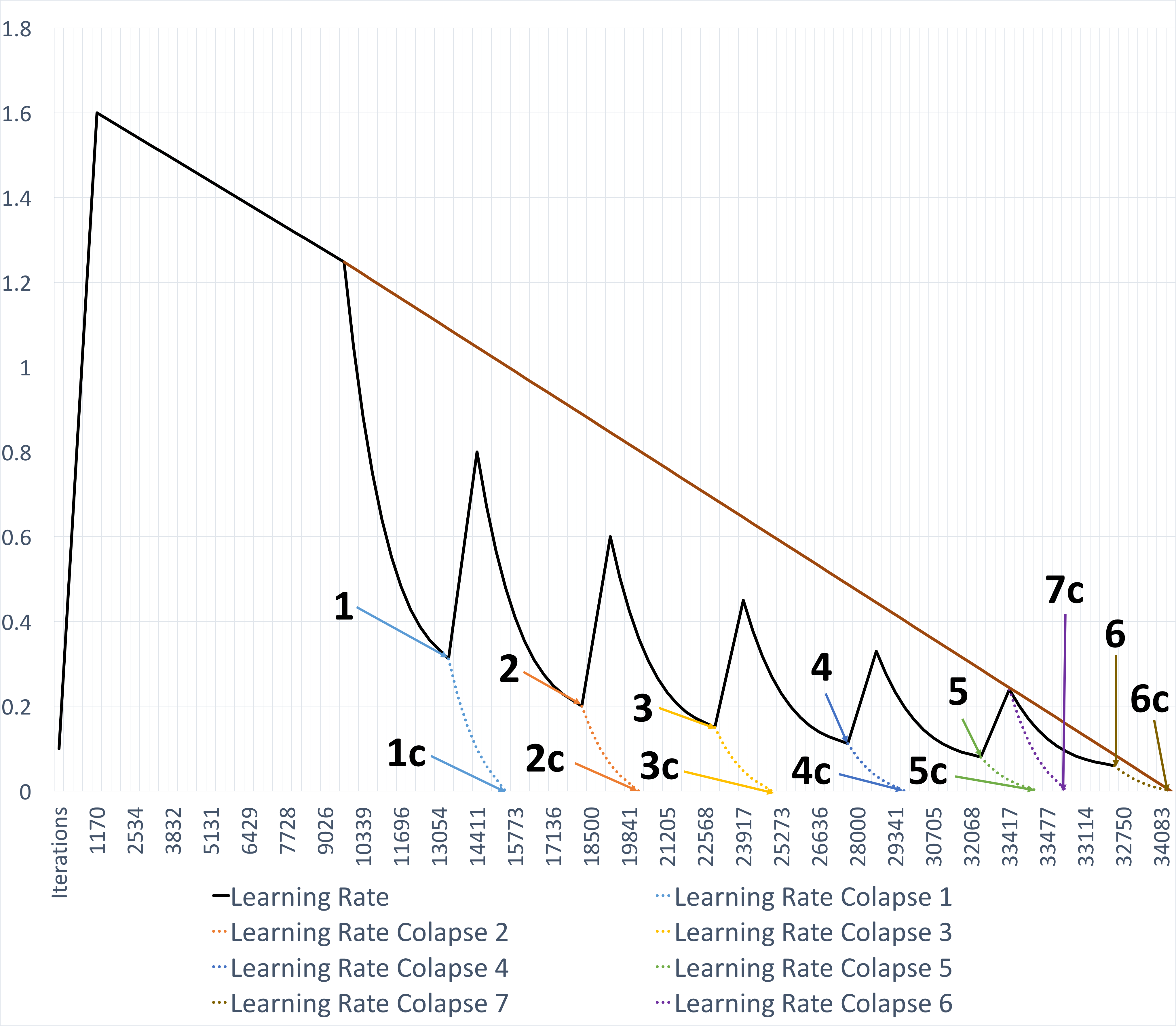}
\caption{Plot of learning rate behaviour and collapse results when obtaining the ensemble snapshots}
\label{fig_collapse_snapshot}
\end{figure}

We have noticed that we can achieve even better validation accuracy, by combining the predictions of ensembles trained using a constant training budget. By using the LR schedule from Figure \ref{fig_collapse_snapshot} we can achieve good any-time performance from a 120-epoch training run. What is different from most training schedules, is that we chose a semi-cyclic learning rate schedule in a fashion similar to \cite{Smith2017}. The training is composed of multiple cycles. The duration and number of cycles controls the accuracy of the resulting model. We have chosen 5 cycles starting at epoch 45. Thus, the first part of the training, until around epoch 30 we linearly decay the learning rate. We then increase the decay slope by employing a power-2 polynomial decay. This is done for another 15 epochs, and at this point , after 45 epochs the learning rate decays to around 22\% of its original value. In order to prevent model overfitting, starting from this point we use a cyclic learning rate schedule, where we increase the learning rate by a factor of 3 in a linear fashion over around 3 epochs (similar to a re-warmup for 3 epochs),  and decay it again with a power of two for the following 12 epochs (The decay is computed such that after 12 epochs, the learning rate is divided by a factor of 4). This results in the first cycle finishing at epoch 60. The same follows for cycles at epoch 75, 90, 105, and 120. An important observation is that this is a hand tuned, empirically derived schedule, in future work we will analyse this behaviour further and encapsulate it in an analytic form. Based on the fact that the final 3 collapses show relatively similar performance, it seems that the amplitude of the cycles needs to be adjusted in order to produce snapshots that converge to different minimas.

Table \ref{table_collapse_snapshot} and Figure \ref{fig_collapse_snapshot} show that already after around 45 epochs, the performance level is above 76\% top-1 accuracy. This grows to a 76.5\% after 75 epochs and ultimately to almost 77\% at 120 epochs. What is even more interesting is that we can create ensembles of these models, what we call \textquotedblleft collapsed ensembles\textquotedblright. Similarly in philosophy to the snapshot ensemble techniques from \cite{Huang2017}, we reuse the same training budget to create an ensemble of snapshots (in Table \ref{table_collapse_snapshot} they are 2c, 3c, 4c, 5c, 6c). Our technique performs better on the ImageNet dataset when ensembling 5 models, achieving a ResNet-50 record 77.5\% accuracy, considering the fact that the ensemble is obtained from a single standard 120 epoch training run.

This schedule is applied for a batch size of 4096, but the same seems to hold for batch sizes of up to 8K/12K.

\begin{table}[!t]
\renewcommand{\arraystretch}{1.3}
\caption{Legend for Figure \ref{fig_collapse_snapshot}}
\label{table_collapse_snapshot}
\centering
\begin{tabular}{|c|c|c|c|}
\hline
Number on figure & Loss & Top-1 \% acc. & Top-5 \% acc. \\
\hline
1 & 1.2835 & 68.33 & 88.71 \\
\hline
1c & 0.974 & 75.50 & 92.83 \\
\hline
2 & 1.1669 & 71.54 & 90.78 \\
\hline
3 & 1.0820 & 73.28 & 91.58 \\
\hline
3c & 0.9468 & 76.50 & 93.24 \\
\hline
4 & 1.0903 & 73.31 & 91.53 \\
\hline
4c & 0.9469 & 76.57 & 93.24 \\
\hline
5 & 1.0786 & 73.89 & 91.97 \\
\hline
5c & 0.9379 & 76.83 & 93.32 \\
\hline
6 & 1.0709 & 74.49 & 92.13 \\
\hline
6c & 0.9541 & 76.81 & 93.32 \\
\hline
7c & 0.9507 & 76.70 & 93.32 \\
\hline
\end{tabular}
\end{table}

Regular large-batch (as well as small-batch) ResNet training uses a weight decay of 0,0001, momentum of 0.9 and linear scaling of the learning rate to accommodate the potentially larger batches. We have empirically noticed that particularly for the large-batch regime it is important to start the training run with a smaller weight decay, and then increase it towards the end of the run, in the "collapsing" phase.  All "collapse" experiments start with a weight decay of 0.00005, thus half of the standard one, and use scale and aspect ratio augmentation. In the 'collapse` phase, we disable data augmentation (while still keeping the random crop), and we double the weight decay to its standard value of 0.0001. In this collapse phase we typically perform between 3-10 epochs, with the larger collapses offering greater accuracy benefits. As described in Section \ref{largest_batch}, gradually increasing the weight decay multiple times during training is advised, particularly for very large batches.

By using the time-per-epoch computed in Table \ref{time_per_epoch}, coupled with the results obtained by the cyclic collapses, we can project a time to a given top-1 accuracy on Stampede2. Note that for these projections we estimate the need of 48 epochs to reach 75.5\%, 64 epochs to reach 76\%, and 78 epochs to reach 76.5\% as demonstrated by the learning rate schedule from Figure \ref{fig_collapse_snapshot} . Table \ref{table_stampede2_accuracy} outlines this projection.

\begin{table}[!t]
\renewcommand{\arraystretch}{1.3}
\caption{Projected time required to reach a given top-1 accuracy on Stampede2 using the collapsed ensemble technique}
\label{table_stampede2_accuracy}
\centering
\begin{tabular}{|c|c|c|c|c|}
\hline
batch size & \# nodes & TT-75.5\%[min] & TT-76\%[min] & TT-76.5\%[min] \\
\hline
4096 & 256 & 82 & 109 & 133  \\
\hline
8192 & 256 & 68 & 90 & 110  \\
\hline
8192 & 512 & 48 & 64 & 78  \\
\hline
10240 & 512 & 44 & 58 & 72  \\
\hline
12288 & 512 & 41 & 55 & 67  \\
\hline
12288 & 768 & 31 & 41 & 49  \\
\hline
\end{tabular}
\end{table}

Using these improved training techniques we can estimate the time require to reach a given validation accuracy level on Skylake as well. Since we've had quite limited budget on MareNostrum 4, we could not perform experiments using these new techniques, but as one may expect, the only difference lays in throughput and scaling efficiency between the systems, and not in statistical efficiency. Thus, based on the throughput data from Table \ref{time_per_epoch_mn4} we can estimate the duration of a ResNet-50 training run to a given validation accuracy level using large-scale Skylake-based nodes. This projection can be found in Table \ref{table_marenostrum4_accuracy}.

\begin{table}[!t]
\renewcommand{\arraystretch}{1.3}
\caption{Projected time required to reach a given top-1 accuracy on MareNostrum 4 using the collapsed ensemble technique}
\label{table_marenostrum4_accuracy}
\centering
\begin{tabular}{|c|c|c|c|c|c|}
\hline
batch size & \# CPUs & TT-75.5\%[min] & TT-76\%[min] & TT-76.5\%[min] \\
\hline
8192 & 512 & 68 & 90 & 110 \\
\hline
9600 & 800 & 50 & 67 & 81 \\
\hline
6400 & 800 & 54 & 72 & 87  \\
\hline
8192 & 1024 & 42 & 56 & 68 \\
\hline
\end{tabular}
\end{table}

\section{Conclusion}

Using a combination of techniques (very large global batch sizes, modified batch normalization, aggressive learning rate schedules, warm-up strategies, weight decay improvements, collapsed ensembles), we have achieved a scalable solution based on Intel's distribution of Caffe that provides state-of-the-art neural network training both in terms of time to trained model and in terms of the achieved accuracy. Moreover, these techniques allowed us to perform various trade-offs between the training time and the accuracy of the resulting model. We have scaled up ResNet-50 training to up to 1536 Knights Landing nodes working with a global batch size of 49152, but still achieving state-of-the-art accuracy in only 28 minutes demonstrating a scaling efficiency of over 80\%. We have also experimented with the Intel Skylake CPUs, and acknowledge it as a viable architecture for training deep neural networks, especially when strong scaling is involved, as the Skylake CPUs sustain small local batches more efficiently.

We presented several techniques allowing us to exceed the baseline accuracy of the ResNet-50 model significantly. By using the proposed collapsed ensemble learning rate schedule technique, we achieved a single-model accuracy of 76.5\% after 75 epochs and an ensemble result of 77.5\%. Similarly in philosophy to the snapshot ensemble techniques from \cite{Huang2017}, we reuse the same training budget to create an ensemble of snapshots. Our technique performs better on the ImageNet dataset when using a 5 model ensemble compared to the one from Huang \emph{et al.}.

As additional future work we are planning to investigate more effective learning rate strategies, such the cosine learning rates also presented in \cite{Loshchilov2016}. Based on our own empirical results and inspired by research such as the Cyclic learning rate \cite{Smith2017} and SGDR \cite{Loshchilov2016}, we believe that as far as SGD based methods are concerned, there is a correlation between a measure of regularity of a dataset, learning rate policy, and maximum global batch size. We plan to investigate a way to automatically learn this relation. More than this, we want to analyze the relationship between width and depth of convolutional networks, total number of inputs, total number of classes, and number of examples per class. More specifically, we will aim at further improving accuracy when using very large batches, but also we plan to use larger-scale datasets, such as the full ImageNet or the Places-365 \cite{Zhou2017} datasets.

All results (models/scripts) presented in this paper will be pushed in the IntelCaffe github repository.



%

\section*{Acknowledgment}

We would like to thank PRACE for awarding access to MareNostrum 4 system based in Spain at BSC. We also want to thank all the consultants from TACC, BSC, and from the VLAB internal cluster for promptly helping us when we had problems to run on the mentioned systems. We also acknowledge our collaboration with Intel under the IPCC framework, and thank Intel for all the support in getting us access to Stampede2 and to the VLAB cluster.

\ifCLASSOPTIONcaptionsoff
  \newpage
\fi



%

%

\begin{IEEEbiography}[{\includegraphics[width=1in,height=1.25in,clip,keepaspectratio]{picture}}]{John Doe}
\blindtext
\end{IEEEbiography}




\end{document}